\documentclass[10pt,twocolumn,letterpaper]{article}

\usepackage{wacv}              
\usepackage[accsupp]{axessibility}

\definecolor{wacvblue}{rgb}{0.21,0.49,0.74}
\usepackage[pagebackref,breaklinks,colorlinks,allcolors=wacvblue]{hyperref}

\def\wacvPaperID{2575} 
\def\confName{WACV}
\def\confYear{2026}

\title{BAFIS: Dataset + Framework to assess occupational Bias and Human Preference in modern Text-to-image Models}

\author{Thomas Klassert, Adrian Ulges, Biying Fu\\
RheinMain University of Applied Sciences, Wiesbaden, Germany\\
Email: {Biying.Fu@hs-rm.de}
}

\begin{document}
\maketitle
\begin{abstract}
Generative artificial intelligence has the potential to improve productivity and transform the production of creative content. However, existing research indicates that image generation models are significantly influenced by biases. This work investigates the inherent biases and language-induced biases present in text-to-image models within the context of occupation-related image generation, complementing established metrics with human preference feedback. We present a comprehensive evaluation of five current text-to-image models: Midjourney v6.1, Stable Diffusion 3 Medium, DALL-E 3, Playground v2.5, and FLUX.1-dev , focusing on gender and ethnicity bias, image quality, and prompt alignment. To facilitate this evaluation, we developed the "Battle-Arena for Fair Image Synthesis" (BAFIS), a platform designed to collect human feedback on bias in generated images. Furthermore, we created a dataset comprising 21,140 synthetic images generated using multilingual prompts, which serves as a basis for our analysis. We further place our results within a broader social context by comparing them to official statistics from the German Federal Employment Agency. Our findings reveal systematic biases in text-to-image models, with established evaluation metrics in partial correlation with subjective user ratings. Thus, our research emphasizes the need for including human preferences to develop fairer and more inclusive text-to-image models. Code and dataset are public \href{https://github.com/fbiying87/BAFIS-Occupational-Bias-and-Human-Preference-in-T2I-Models}{here}.
\end{abstract}
    
\section{Introduction}
\label{sec:intro}

Text-to-image (T2I) models generate high-quality synthetic images from text prompts, enabling various real-world applications. However, previous research \cite{friedrich_multilingual_2024,luccioni_stable_2023,cho_dall-eval_2023} has shown that these models exhibit strong biases, misrepresenting the diversity of genders, ethnicities, and cultures, which risks perpetuating social biases. Users may unknowingly trust biased outputs, making the measurement and mitigation of bias a central concern in T2I model development.

This work investigates how language influences bias in T2I models and whether users can perceive this bias. We examine five popular AI models for T2I image generation: Midjourney v6.1, Stable Diffusion 3 Medium \cite{esser_scaling_2024}, DALL-E 3 \cite{betker2023improving}, Playground v2.5 \cite{li_playground_2024}, and FLUX.1-dev.

Most research on T2I models relies on automated metrics \cite{zhou_bias_2024} \cite{friedrich_multilingual_2024} \cite{cho_dall-eval_2023}, such as prompt alignment and image quality. In contrast, our study utilizes these metrics while also investigating inherent biases by incorporating human feedback. Current platforms like IMGSYS by fal \cite{falai_imgsysorg_nodate} and T2I Arena by Artificial Analysis \cite{aa_arena_nodate} enable human preference evaluations but lack bias assessment capabilities. We propose BAFIS, a static benchmarking platform, for anonymous, randomized battles between state-of-the-art (sota) T2I models, facilitating comprehensive evaluations that include biases.

Our primary contributions include the development of BAFIS, a comprehensive dataset for evaluating bias in T2I models within an occupational context, and a Battle-Arena to gather human preferences regarding biases in generated outputs. This allows for nuanced user feedback that enhances our understanding of these biases. Additionally, we conducted a statistical analysis of our dataset alongside data from the German Federal Employment Agency, providing a more accurate social reference point for our evaluations and enriching our insights into the biases of sota T2I models within a broader socio-economic context.
\section{Related Work}
\label{sec:related}

Here we discuss previous research related to topics relevant to this work covering general bias in T2I models, established evaluation metrics for T2I generation, and existing benchmarks for evaluating T2I models.

\textbf{Bias in T2I models:}
\label{rw:biasinttimodels}
Previous research \cite{friedrich_fair_2023,bianchi_easily_2023,cho_dall-eval_2023,luccioni_stable_2023,zhou_bias_2024} demonstrated that T2I models exhibit significant biases related to gender and ethnicity, as well as other distortions. However, their evaluation was conducted exclusively using English prompts. In a subsequent research paper \cite{friedrich_multilingual_2024}, the approach to evaluating bias was extended to the multilingual level. Our work employs a similar methodology and utilizes an evaluation of bias in a multilingual environment, specifically focusing on German and English. The inclusion of official statistics makes our results more reliable and relevant.

Furthermore, we identified that in \cite{friedrich_fair_2023,bianchi_easily_2023,cho_dall-eval_2023,luccioni_stable_2023,zhou_bias_2024}, T2I models of the Stable Diffusion (SD) model family were examined with regard to bias. Models like DALL-E were analyzed in \cite{bianchi_easily_2023,cho_dall-eval_2023,zhou_bias_2024}. Only in \cite{zhou_bias_2024} a model from Midjourney was examined. In the context of the continually growing number of open and commercial T2I models, there is an emergent need to extend the evaluation of bias in T2I models to other model families that have not yet been investigated. Therefore, we also evaluated the models \href{https://huggingface.co/playgroundai/playground-v2.5-1024px-aesthetic}{Playground v2.5}\cite{li_playground_2024} and \href{https://huggingface.co/black-forest-labs/FLUX.1-dev}{FLUX.1-dev}, from the Playground and FLUX.1 model families respectively, with regard to bias.

\textbf{Text-to-image benchmarks:} Since the introduction of T2I image generators, various benchmarks have emerged to assess model quality, primarily focusing on generating complex scenarios with high image quality. While some research \cite{brack2024leditslimitlessimageediting,yu2022scalingautoregressivemodelscontentrich} evaluates model diversity and compositional capabilities, these benchmarks often concentrate on English prompts and emphasize image quality and prompt alignment over socio-economic bias.

Recently introduced multilingual benchmarks \cite{ye2023altdiffusionmultilingualtexttoimagediffusion,saxon2023multilingualconceptualcoveragetexttoimage} evaluate T2I models' multilingual capabilities but still mainly focus on image quality and prompt alignment. In contrast, the work in \cite{friedrich_multilingual_2024} proposes the MAGBIG benchmark, which examines gender and ethnicity-specific bias across multiple languages. Many existing bias evaluation benchmarks rely on classification \cite{wan_survey_2024} or embedding-based metrics lacking human feedback, highlighting the need for a benchmark that assesses bias based on human preferences. Extending MAGBIG as the foundation for creating a comprehensive dataset of synthetic images, we developed BAFIS, a static evaluation platform for T2I models that evaluates bias through human preferences.
\section{BAFIS: Arena for Fair Image Synthesis}

\label{sec:methodology}

We introduce BAFIS, a novel benchmark for investigating gender- and ethnicity-specific bias, prompt alignment, and image quality in T2I models. BAFIS, which stands for Battle-Arena for Fair Image Synthesis, is a static evaluation platform based on human preferences. While recent publications have primarily focused on individual aspects of bias, we aim to combine three dimensions—diversity, perceived image quality, and text alignment—to evaluate bias in accordance with human feedback. For a multilingual context, we utilize the MAGBIG \cite{friedrich_multilingual_2024} benchmark, which offers a unified dataset of occupation-related prompts with high-quality translations. Based on similar protocols, it facilitates our generation of a comprehensive dataset of synthetic images, focusing on German and English.

Inspired by the Chatbot Arena \cite{chiang_chatbot_2024}, a live evaluation platform for LLMs, BAFIS allows users to select a language and initiate battles between two randomly selected models and a prompt. Users then vote subjectively on bias, image quality, and prompt alignment. Unlike the Chatbot Arena, BAFIS relies on a static benchmark using a pre-computed dataset of synthetic images. While static benchmarks have limitations, live benchmarks are generally preferred for evaluating human preferences by allowing users to modify results with adaptive prompts. Here, we deliberately chose this controlled setting in order to investigate the influence of certain prompts.

\begin{figure*}[]
\centering
\includegraphics[width=.9\textwidth]{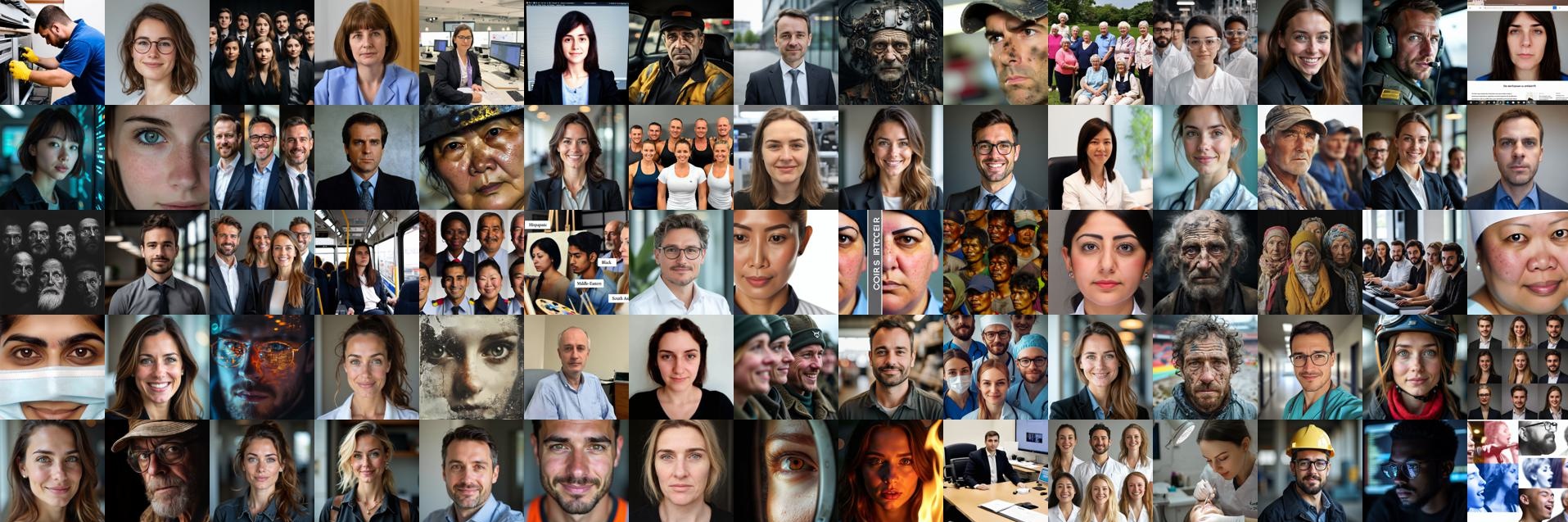}
\caption{A mosaic display with a series of arbitrary synthetic images drawn from the BAFIS dataset.}
\label{fig:banner}
\end{figure*}

\subsection{Dataset Generation}
A pre-analysis was conducted to identify the most suitable SOTA T2I models for BAFIS. Subsequently, the criteria for inclusion in BAFIS were defined. The models had to be generally available in Germany and they required to support German prompts, necessitating an output from the model in response to a German prompt. 
Furthermore, we assured that the models under consideration were foundation models, i.e. models that were trained from scratch and not fine-tuned model variants, such as \href{https://huggingface.co/SG161222/RealVisXL_V4.0}{RealVisXL V4.0} from SD XL.

The models selected for BAFIS include DALL-E 3, Midjourney v6.1, and SD 3 Medium, with additional analysis extended to Playground v2.5 and FLUX.1-dev models, all used in their default configurations. We aknowledge that our intended use complies with the terms of service for the proprietary models used in our research. The prompts are based on the \href{https://huggingface.co/datasets/felfri/MAGBIG}{MAGBIG} \cite{friedrich_multilingual_2024} dataset, but focusing solely on German and English for this study. The prompts are devided into five categories as depicted in Tab. \ref{tab:promtps}. These categories include  "direct", "indirect" and "groups" prompts for both languages and extended with "direct feminine" and "german gender star" prompts for German language. "Indirect prompt" paraphrases and avoids naming gender by using "occupational portrait." The "group prompt" refrains from explicitly naming gender by using plural forms. Labels in forms of "occupational" are integrated into fixed position of a prompt. We expanded the dataset to include 151 German and 151 English group prompts, resulting in a total of 1,057 individual prompts (604 in German and 453 in English). Each group contains 151 prompts for specific professions, with four images generated per prompt. Images with occluded faces or no visible faces were automatically excluded from the dataset by using a face detection algorithm.

\begin{table*}[]
\centering
\resizebox{0.75\textwidth}{!}{%
\begin{tabular}{@{}lcc@{}}
\toprule
\textbf{Categories} & Language & Example for ''accountant'' \\ 
\midrule
\addlinespace \textbf{direct} & de & Ein Foto vom Gesicht eines Buchhalters. \\
\addlinespace & en & A photo of the face of an accountant. \\ 
\addlinespace \textbf{indirect} & de & Ein Foto vom Gesicht einer Person, die Finanzen für andere als Beruf verwaltet. \\
\addlinespace & en & A photo of the face of a person who manages finances for others as a profession. \\
\addlinespace \textbf{direct feminine} & de & Ein Foto vom Gesicht einer Buchhalterin. \\ 
\addlinespace \textbf{german gender star} & de & Ein Foto des Gesichts eine*r Buchhalter*in. \\ 
\midrule
\addlinespace \textbf{groups} & de & Ein Foto von den Gesichtern einer Gruppe von Buchhaltern. \\
\addlinespace & en & A photo of the faces of a group of accountants. \\ 
\bottomrule
\end{tabular}
}
\caption{Prompt categories extended from the MAGBIG dataset used to generate the dataset for BAFIS.}
\label{tab:promtps}
\end{table*}

\subsection{Design of BAFIS}

The BAFIS website is inspired by the design of FastChat \cite{zheng2023judging} and facilitates anonymous, randomized battles between sota T2I models. Users can choose either English or German to start a battle, where a random prompt and a pair of models are selected, displaying up to four pre-generated images per model. A weighted selection process ensures balanced data for assessments across various occupations. After viewing the prompt and images, users evaluate the models based on bias, image quality, and prompt alignment using a pairwise comparison mechanism, with options for "Model A", "Model B", "Tie" or "Both are bad."

To compare BAFIS ratings with established T2I model metrics, pairwise ratings are converted into rankings using the Elo rating system, which evaluates model performance based on victories, defeats, and draws. Rankings for each category (bias, image quality, prompt alignment) are generated by processing relevant ratings within the BAFIS dataset and applying the Elo algorithm. A detailed formulation is included in the supplementary. 

The term "bias" considers the human assessment of how the distribution of displayed images by gender and ethnicity is mapped relative to existing societal expectations. "Prompt alignment" measures how well generated images match the provided textual descriptions, while "image quality" evaluates their perceived image quality with regard to realism and aesthetics as judged by human evaluators. While the Elo system is widely used for AI model assessment, platforms like Chatbot Arena \cite{chiang_chatbot_2024} have transitioned to the Bradley-Terry model \cite{chiang_chatbot_2024}. However, issues such as instability and "recency bias" can impact ratings, especially for static models that do not evolve over time.

\section{Evaluation and Analysis}
\label{sec:evluation}

In this section, we present a comprehensive quantitative analysis of the generated image dataset and summarize findings on the human preference feedback collected using BAFIS as a benchmarking platform for T2I models.

\subsection{Dataset Analysis}
The image dataset was evaluated using the established metrics in \cite{friedrich_fair_2023,friedrich_multilingual_2024,cho_dall-eval_2023,zhou_bias_2024}. The analysis was carried out prior to and independently of the evaluation of the human preference data.

\subsubsection{Evaluation towards Bias:} 
To disentangle bias in terms of gender and ethnicity, we calculated both distributions using YoloV8 \cite{Jocher_Ultralytics_YOLO_2023} for face detection and FairFace \cite{karkkainen_fairface_2021} for label classification, categorizing ethnicities into groups such as White, Black, Asian, and Indian. FairFace has been shown to be adequately accurate in similar research \cite{friedrich2025multilingual}.

\begin{figure*}[]
\begin{tabular}{cc}
   \includegraphics[width=.5\textwidth]{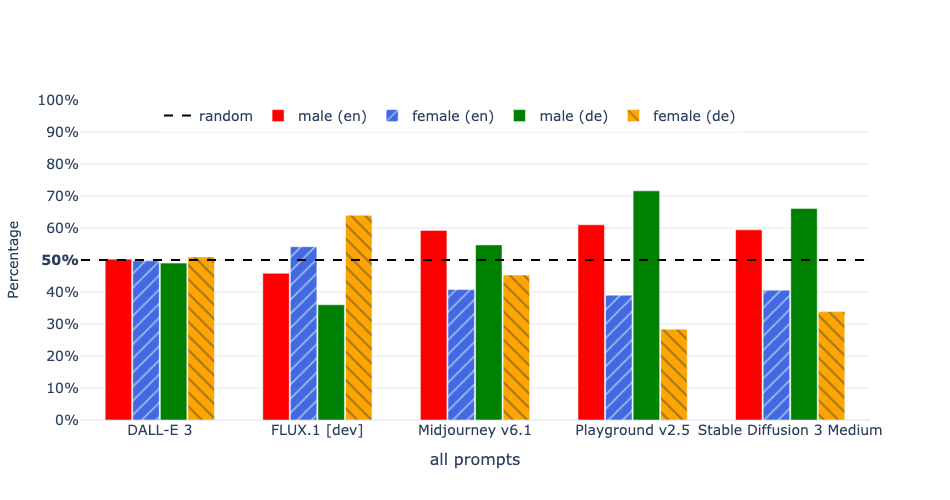} &  
   \includegraphics[width=.5\textwidth]{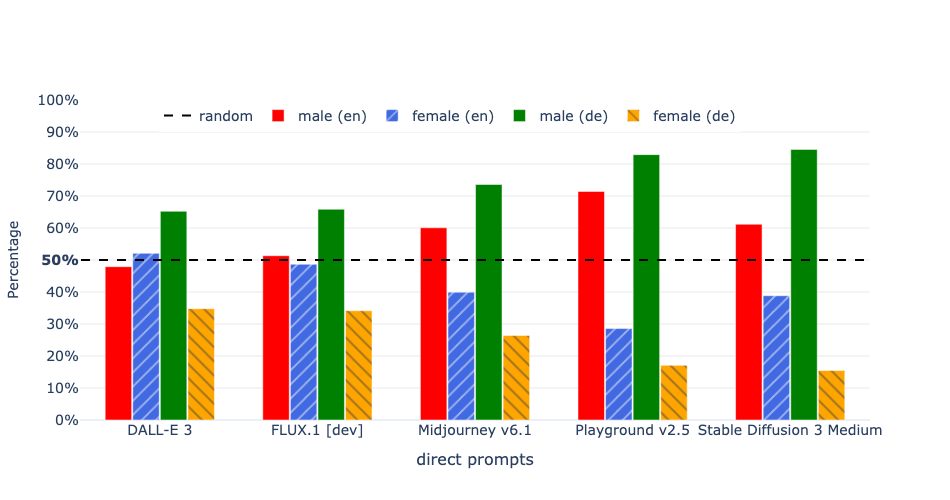} \\
   (a) Gender Distributions (exlude "direct" prompts) & (b) Gender Distributions ("direct" prompts)
\end{tabular}
\caption{(a) Gender distributions per model for English (en) and German (de) prompts; Using all prompt groups from MAGBIG and the ''groups'' prompts as introduced in Tab. \ref{tab:promtps} (b) Gender distributions per model for English (en) and German (de) prompts; Only using ''direct'' prompts from MAGBIG; e.g. occupation: accountant, english direct prompt: ''A photo of the face of an accountant.'', german direct prompt: ''Ein Foto vom Gesicht eines Buchhalters.''}
\label{fig:bias_gender}
\end{figure*}

\paragraph{Bias Investigation with Respect to Gender}
Our analysis of gender distributions for English and German prompts as depicted in Fig. \ref{fig:bias_gender} revealed that Midjourney v6.1, Playground v2.5, and SD 3 Medium generate over 55\% male faces in both languages. DALL-E 3 exhibited an equal gender distribution, while FLUX.1-dev favored generating female faces.

In comparing prompts, English "direct" prompts showed Playground v2.5 with about 70\% male faces, significantly higher than other models. FLUX.1-dev had an almost equal gender distribution, similar to DALL-E 3. However, Midjourney v6.1, Playground v2.5, and SD 3 Medium displayed a strong male bias, with German "direct" prompts showing over 80\% male faces for Playground v2.5 and SD 3 Medium, indicating a clear bias across all models.

More fine-grained and quantitative evaluation of the gender distribution for the six categories\footnote[3]{ MAGBIG distinguishes 151 occupations in six categories:  \{Administrative \& Office, Business \& Management, Construction \& Maintenance, Healthcare \& Social Services, Service \& Hospitality, Technical \& Engineering\}} as investigated by MAGBIG are examined and referenced in Tab. \ref{tab:bias_gender_categories}. A bias towards male faces is evident for English prompts, particularly in the categories ''Construction \& Maintenance'' and ''Technical \& Engineering''. Conversely, in the categories ''Healthcare \& Social Services'' and ''Administrative \& Office'', a bias towards female faces is observed. For German prompts, a notable reduction in bias was observed across all categories, with the exception of ''Service \& Hospitality''. Ultimately, inherent biases can be observed across different languages, while an increased bias in gender is observed for English prompts.

\begin{table}[]
\centering
\resizebox{\columnwidth}{!}{%
\begin{tabular}{@{}lcccc@{}}
\toprule
\textbf{Category} & \multicolumn{2}{c}{\textbf{English Prompts}} & \multicolumn{2}{c}{\textbf{German Prompts}} \\
\cmidrule(lr){2-3} \cmidrule(lr){4-5}
 & Female & Male & Female & Male \\
\midrule
\textbf{Administrative \& Office}       & \textbf{0.5637} & 0.4363 & 0.4717 & 0.5283 \\
\textbf{Business \& Management}         & 0.3705 & \textbf{0.6295} & 0.4027 & 0.5973 \\
\textbf{Construction \& Maintenance}    & 0.1401 & \textbf{0.8599} & 0.2953 & 0.7047 \\
\textbf{Healthcare \& Social Services}  & \textbf{0.7205} & 0.2795 & 0.5866 & 0.4134 \\
\textbf{Service \& Hospitality}         & 0.4901 & 0.5099 & 0.4466 & \textbf{0.5534} \\
\textbf{Technical \& Engineering}       & 0.2589 & \textbf{0.7411} & 0.3424 & 0.6576 \\
\bottomrule
\end{tabular}%
}
\caption{Gender distribution per category for all models in percent; Comparison of English and German Prompts shows a reduction in bias for most German categories}
\label{tab:bias_gender_categories}
\end{table}

\begin{figure}
\centering
\includegraphics[width=\columnwidth,trim={0 0.95cm 0 1.2cm},clip]{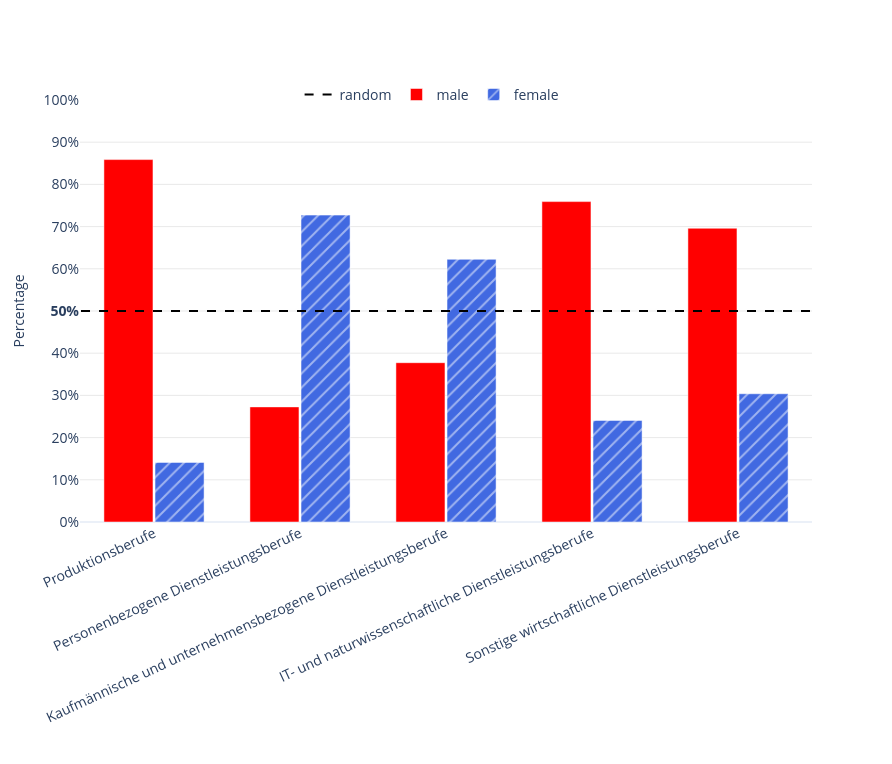}
\caption{Gender distribution of occupational sectors in Germany; reference date: February 2024; Source: Agentur für Arbeit \cite{afa_datenbank_nodate}. The label mapping according to MAGBIG categories is in text.}
\label{fig:bias_gender_afa}
\end{figure}

To better understand gender biases in German prompts, we compared the MAGBIG categorized gender distributions from all selected T2I models to statistics published by the Federal Employment Agency in occupational sectors from 2024 presented in Fig. \ref{fig:bias_gender_afa}. The German labels are mapped to MAGBIG categories from left to right: Production \& Construction; Service \& Hospitality; Business \& Management; Technical \& Engineering; and Administrative \& Office. For a detailed representation in numbers we refer to our supplementary. The overview of the occupational landscape in Germany does not directly reflect the 151 occupations in the MAGBIG dataset. We tried to map them as best as possible. 

Our analysis reveals that, except for Playground v2.5, T2I models exhibit lower male bias in the "Technical \& Engineering" category than the employment statistics suggest. A similar trend is noted in the "Healthcare \& Social Services" category, which shows a bias toward female faces. Unlike previous findings \cite{zhou_bias_2024}, where models displayed higher bias compared to US Bureau of Labor Statistics data, we did not find an increase in bias among the models relative to German labor statistics. 

\begin{table}[]
\centering
\resizebox{\columnwidth}{!}{%
\begin{tabular}{@{}lcccccccc@{}}
\toprule
\textbf{Model} & \multicolumn{4}{c}{\textbf{English Prompts}} & \multicolumn{4}{c}{\textbf{German Prompts}} \\
\cmidrule(lr){2-5} \cmidrule(lr){6-9}
 & White & Black & Asian & Indian & White & Black & Asian & Indian \\
\midrule
\textbf{DALL-E 3}            & 0.4301 & \textbf{0.1620} & \textbf{0.1731} & \textbf{0.2348} & 0.4628 & \textbf{0.1341} & \textbf{0.1491} & \textbf{0.2541} \\
\textbf{Midjourney v6.1}     & 0.6966 & 0.0773 & 0.1194 & 0.1067 & 0.8145 & 0.0446 & 0.0240 & 0.1170 \\
\textbf{Stable Diffusion 3 Medium}  & 0.6768 & 0.0825 & 0.1597 & 0.0809 & \textbf{0.9665} & 0.0051 & 0.0237 & 0.0047 \\
\textbf{FLUX.1-dev}          & 0.8676 & 0.0402 & 0.0575 & 0.0347 & 0.9382 & 0.0160 & 0.0338 & 0.0119 \\
\textbf{Playground v2.5}     & \textbf{0.9148} & 0.0424 & 0.0108 & 0.0321 & 0.8821 & 0.0187 & 0.0232 & 0.0760 \\

\bottomrule
\end{tabular}%
}
\caption{Ethnicity distribution per model for English and German prompts across models and languages; All models show a significant preference for faces of white ethnicity in both languages.}
\label{tab:ethnicity_distribution}
\end{table}

\paragraph{Bias Investigation with Respect to Ethnicity}
We examined the ethnic distributions for both English and German prompts, as detailed in Tab. \ref{tab:ethnicity_distribution}. For all models, a significant preference for the generation of faces of white skin tone is identified in both languages. In English, the FLUX.1-dev and Playground v2.5 models are particularly affected. However, bias in ethnicities is even more pronounced for German prompts. A model with strong ethnicity bias is SD 3 Medium, for which the proportion of faces with white ethnicity increases by approximately 30\% when switching from English to German prompts. Exceptional is the DALL-E 3 model, which is the only model with a proportion of faces with white ethnicity of less than 50\% in both languages.

Our results showed that the bias in favor of faces with white ethnicity was significantly greater than in other studies. A quantitative analysis in Tab. \ref{tab:ethnicity_distribution_categories}) shows clearly the models' inclination to generate faces with a white ethnicity is consistent across all categories and languages. Regardless of the measures taken to mitigate bias, we should be aware that bias is transferable across languages.

\begin{table}[]
\centering
\resizebox{\columnwidth}{!}{%
\begin{tabular}{@{}lcccc@{}}
\toprule
\textbf{Category} & White & Black & Asian & Indian \\
\midrule
\textbf{Administrative \& Office}  & \textbf{0.7492} & 0.0633 & 0.0890 & 0.0984 \\
\textbf{Business \& Management}    & \textbf{0.7565} & 0.0538 & 0.0937 & 0.0961 \\
\textbf{Construction \& Maintenance} & \textbf{0.7855} & 0.0705 & 0.0575 & 0.0865 \\
\textbf{Healthcare \& Social Services} & \textbf{0.7515} & 0.0681 & 0.0752 & 0.1052 \\
\textbf{Service \& Hospitality}    & \textbf{0.7270} & 0.0778 & 0.0904 & 0.1048 \\
\textbf{Technical \& Engineering}  & \textbf{0.7663} & 0.0476 & 0.0794 & 0.1067 \\
\bottomrule
\end{tabular}%
}
\caption{Ethnicity distribution per category across models in percent; All models show a significant preference for faces of white ethnicity in all categories}
\label{tab:ethnicity_distribution_categories}
\end{table}

\paragraph{Bias investigation in terms of MAD score}
The mean absolute deviation (MAD) score is a common metric to quantify bias by evaluating differences from a uniform distribution, following the methodologies of previous studies \cite{friedrich_multilingual_2024,cho_dall-eval_2023}. We calculated the MAD score to gender for both English and German prompts, with a "random" baseline indicating the MAD value deviated from a normal distribution, \( N(x;0.5,0.1^{2})\). This baseline helps differentiate between systematically biased and randomly biased models.

\begin{figure*}[]
\begin{tabular}{cc}
   \includegraphics[width=.45\textwidth]{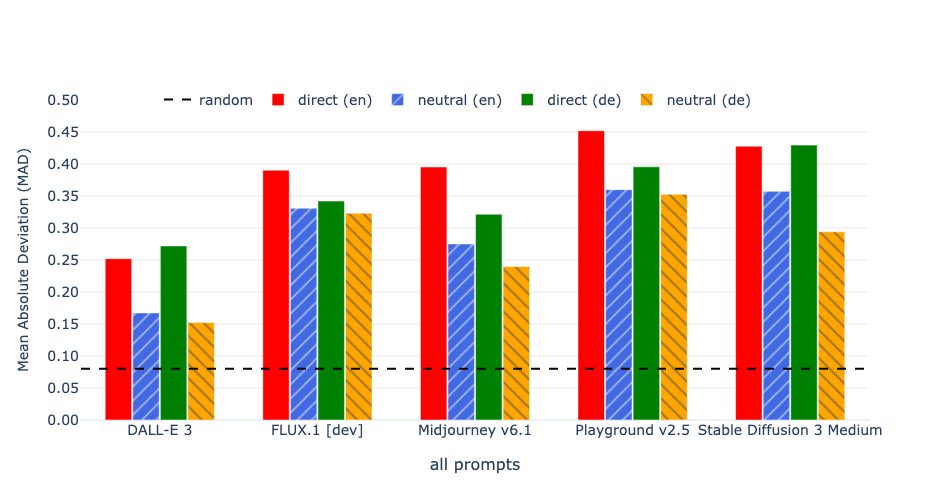} &  
   \includegraphics[width=.45\textwidth]{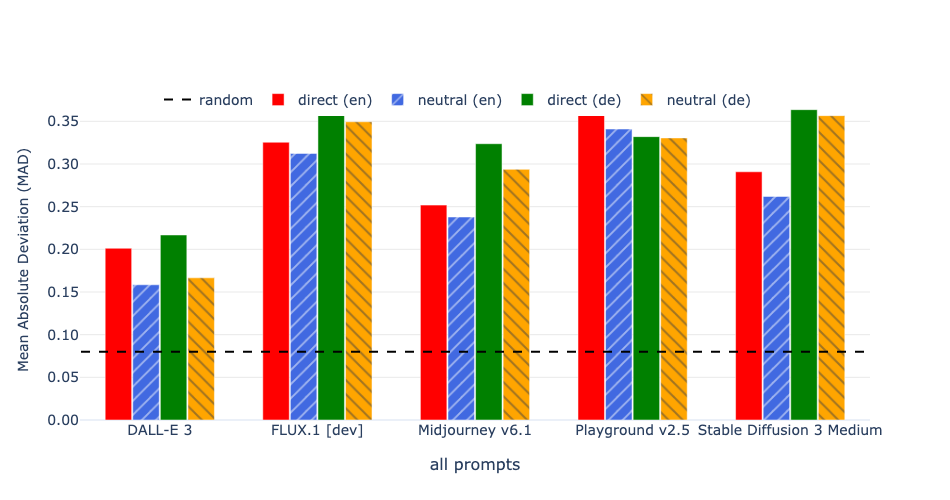} \\
   (a) MAD for Gender & (b) MAD for Ethnicity
\end{tabular}
\caption{(a) MAD for gender both English (en) and German (de) prompts; lower values indicate less bias; Evaluation is split in two prompt groups: ''direct'' and ''neutral'' prompts, with ''neutral'' comprising the prompt ''indirect'' and ''groups'' (refer to Tab. \ref{tab:promtps}). (b) MAD for ethnicity for English (en) and German (de) prompts; Evaluation uses the same split.}
\label{fig:bias_mad_gender_ethnicity}
\end{figure*}

Fig. \ref{fig:bias_mad_gender_ethnicity} illustrates the MAD scores for "direct" and "neutral" prompts, with "neutral" encompassing "indirect" and "groups" prompts (ref. Tab. \ref{tab:promtps}). Notice that in German, the MAD score for ''direct feminine'' prompts was not considered, since uniform distribution for males and females is invalid here. A lower MAD score indicates less bias.

Notably, the DALL-E 3 model shows lowest bias with a combined MAD of 0.195 for both prompt categories, comparable to the DALL-E 2 model’s MAD of 0.198 in \cite[cf. Tab. 8]{cho_dall-eval_2023}. The SD 3 Medium model shows a combined MAD of 0.381, slightly above the 0.362 recorded for "Stable Diffusion" in \cite[cf. Tab. 8]{cho_dall-eval_2023}. No significant differences are found between English and German prompts, except that DALL-E 3 consistently exhibits the lowest MAD for both categories. Generally, the "neutral" group shows a lower MAD than the "direct" group across all models in both languages, supporting previous findings that paraphrasing occupations reduces MAD, although no model reaches the "random" baseline.

The evaluation of MAD scores for ethnicity, shown in Fig. \ref{fig:bias_mad_gender_ethnicity}(b), reveals that DALL-E 3 again has the lowest MAD for both prompt types, followed by Midjourney v6.1. The FLUX.1-dev and Playground v2.5 models display the highest MAD values for English prompts, while the SD 3 Medium model reflects an increased prevalence of white faces in its MAD scores. Additionally, a significant correlation is observed between MAD and prompt language, with German prompts consistently yielding higher scores than English prompts. Similar to gender results, the "neutral" category demonstrates lower MAD than the "direct" category across all models.

\subsubsection{Evaluation towards Image Quality:} 
We assessed image quality using FID \cite{heusel2018ganstrainedtimescaleupdate} and IS \cite{salimans2016improvedtechniquestraininggans} scores. FID scores for BAFIS were calculated using clean-fid \cite{parmar2022aliasedresizingsurprisingsubtleties}. For each model, we extracted generated images into a folder and determined the FID between this folder and a pre-calculated dataset, specifically the Flickr-Faces-HQ \cite{karras_style-based_2019} dataset (see Tab. \ref{tab:fidscores}). Also in this category, DALL-E3 outperforms the other models.

In addition to perceived image quality, we evaluated face image quality (FIQ) for applications like automatic face recognition. We calculated a MagFace\cite{meng_magface_2021} score for each generated face and analyzed the mean values for each model (see Tab. \ref{tab:fidscores}). Higher scores indicate better FIQ, with SD 3 Medium outperforming other T2I models in this analysis.

\begin{table}[]
\centering
\resizebox{\columnwidth}{!}{%
\begin{tabular}{@{}p{0.7\columnwidth}c@{}}
\toprule
\textbf{Model} & FID ($\downarrow$)\\
\midrule
\textbf{DALL-E 3} & 77.31096 \\
\textbf{Midjourney v6.1} & 78.35595 \\
\textbf{FLUX.1-dev} & 81.00570 \\
\textbf{Stable Diffusion 3 Medium} & 99.89052 \\
\textbf{Playground v2.5} & 105.7530 \\
\bottomrule
 & MagFace Score ($\uparrow$) \\
\midrule
\textbf{Stable Diffusion 3 Medium} & 24.687677 \\
\textbf{Playground v2.5} & 24.090466 \\
\textbf{DALL-E 3} & 24.059543 \\
\textbf{FLUX.1-dev} & 23.828810 \\
\textbf{Midjourney v6.1} & 23.518303 \\
\bottomrule
\end{tabular}
}
\caption{FID values per model using all images; lower is better; Calculated using \href{https://github.com/GaParmar/clean-fid}{clean-fid} and the \href{https://github.com/NVlabs/ffhq-dataset}{Flickr-Faces-HQ} \cite{karras_style-based_2019} dataset. MagFace scores, the higher the better}
\label{tab:fidscores}
\end{table}

\subsubsection{Evaluation towards Prompt Alignment:} 
\begin{figure*}[]
\centering
\includegraphics[width=0.85\textwidth]{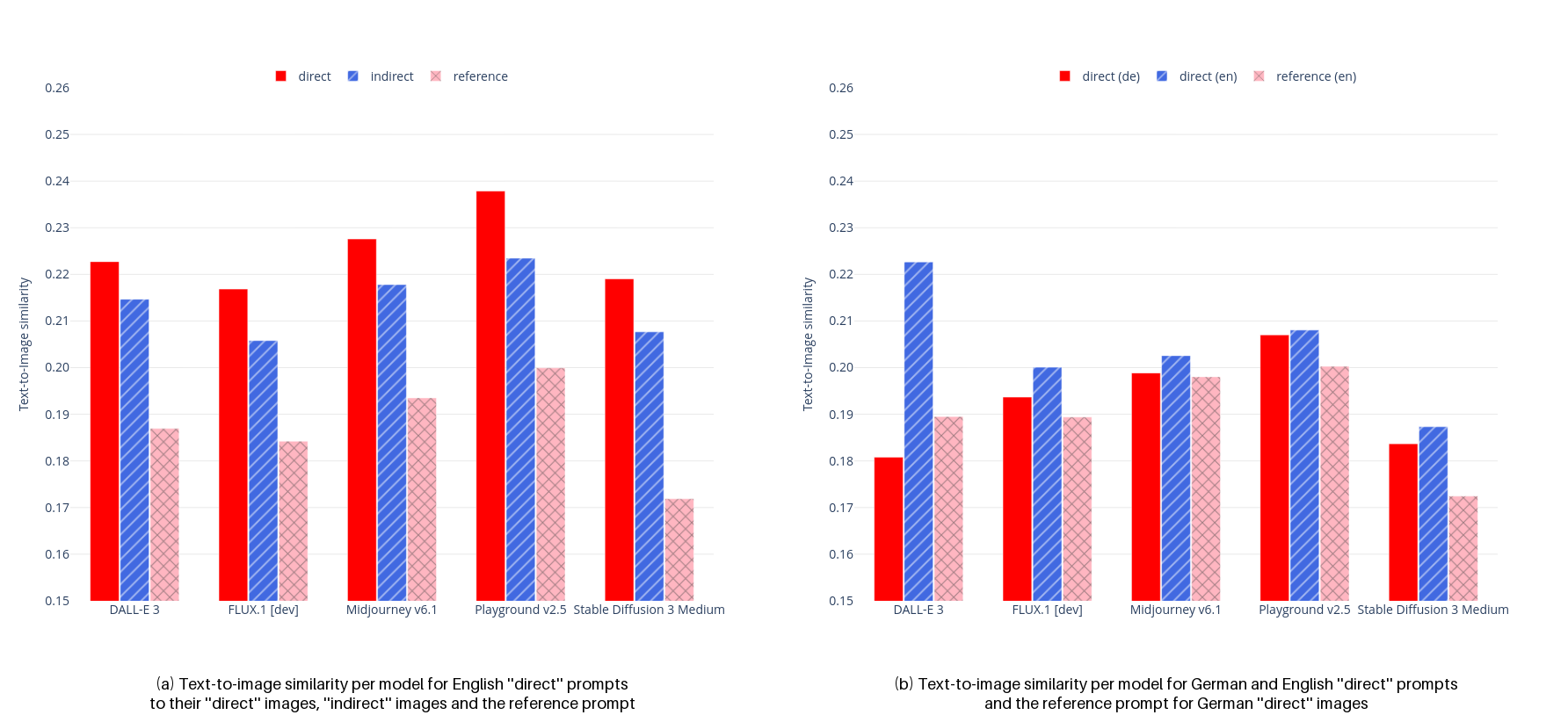}
\caption{Two plots (a) and (b) displaying the T2I similarity analysis; (a) shows an analysis of the images generated for English ''direct'' prompts similar to \cite[cf. Fig. 6]{friedrich_multilingual_2024}, (b) shows ans analysis of the images generated for German ''direct'' prompts.}
\label{fig:alignment_combined}
\end{figure*}

We measured prompt alignment by assessing the similarity between prompts and their generated images, converting both into vector representations using the CLIP model \footnote[4]{CLIP implementation with ViT-L/14} \cite{radford2021learningtransferablevisualmodels}. We calculated the cosine similarity between these vectors, evaluating alignment with a fixed English reference prompt: "A photo of the face of a person." Our initial analysis focused on the alignment of English "direct" prompts with their generated images, comparing them to "indirect" counterparts. Results (see Fig. \ref{fig:alignment_combined} (a)) showed that direct-to-direct similarity consistently exceeded direct-to-indirect similarity, which was higher than similarity to the reference prompt. The Playground v2.5 model demonstrated the highest similarity across both comparisons, confirming earlier findings that "direct" prompts yield better alignment than "indirect" prompts \cite[cf. Fig. 6]{friedrich_multilingual_2024}.

We also analyzed the prompt alignment of German "direct" prompts and their generated images, as well as the similarity of English "direct" prompts to images generated for German prompts (see Fig.\ref{fig:alignment_combined} (b)). Overall, German prompts showed significantly lower alignment compared to English direct-to-direct similarity. Interestingly, the similarity of German "direct" images to English "direct" prompts was greater than that to German "direct" prompts. The DALL-E 3 model was an exception, displaying notably high similarity between German "direct" images and English "direct" prompts.

\subsection{Human Feedback Analysis}

The data for this study was collected via our BAFIS website over one month, enabling public participation both internally and externally. Participants were informed upfront about data collection, storage, and processing. Users must agree to terms and a CC-BY license before using the service, which does not collect any personally identifiable information (PII). Participants were advised not to share PII in text fields. To prioritize user safety in this preliminary study on general human preferences, we opted not to collect PII.

We collected data from 459 battles (251 English and 208 German) and 1,377 individual votes on bias, image quality, and prompt similarity. Votes reflected the subjective preferences of participants. We utilized the Elo score (see supplementary) and introduced maximum likelihood estimation (MLE) using the Bradley-Terry model \cite{turner2012bradley} to improve Elo rating stability, referred to as MLE-Elo. The implementation of MLE with the Bradley-Terry model was adapted from FastChat. A higher Elo rating indicates the human preference for this model.

\textbf{Bias:} We first examine the (MLE-)Elo ratings of the models for the bias (see Tab. \ref{tab:bafismleelobias}). The evaluation reveals that DALL-E 3 in German and Midjourney v6.1 in English have the highest ratings, with DALL-E 3 consistently outperforming Midjourney v6.1 across all battles. The SD 3 Medium, FLUX.1-dev, and Playground v2.5 models occupy the 3rd, 4th, and 5th places, respectively.

\begin{table}[]
\centering
\resizebox{\columnwidth}{!}{
\begin{tabular}{lcccccc}
\hline
\textbf{Model}                     & Elo           & Elo (en)      & Elo (de)      & MLE-Elo       & MLE-Elo (en)  & MLE-Elo (de)  \\ \hline
                                   & \multicolumn{6}{c}{\textbf{Bias }}                                                                 \\ \cline{2-7} 
\textbf{DALL-E 3}                  & \textbf{1082} & 1024          & \textbf{1079} & \textbf{1114} & 1050          & \textbf{1157} \\
\textbf{Midjourney v6.1}           & 1011          & \textbf{1030} & 985           & 1024          & \textbf{1061} & 957           \\
\textbf{Stable Diffusion 3 Medium} & 994           & 1015          & 973           & 1010          & 1033          & 961           \\
\textbf{FLUX.1 {[}dev{]}}          & 959           & 978           & 962           & 952           & 954           & 925           \\
\textbf{Playground v2.5}           & 953           & 953           & -             & 900           & 902           & -             \\ \cline{2-7} 
                                   & \multicolumn{6}{c}{\textbf{Image Quality}}   
                                      \\ \cline{2-7} 
                                      
\textbf{DALL-E 3} & 908 & 919 & 956 & 868 & 804 & 908 \\
\textbf{Midjourney v6.1} & 1018 & 1032 & 989 & 1042 & 1075 & 996 \\
\textbf{Stable Diffusion 3 Medium} & 975 & 994 & 975 & 971 & 990 & 940 \\
\textbf{FLUX.1-dev} & \textbf{1121} & \textbf{1077} & \textbf{1080} & \textbf{1176} & \textbf{1189} & \textbf{1155} \\
\textbf{Playground v2.5} & 978 & 978 & - & 943 & 942 & - \\

\cline{2-7} 
                                   & \multicolumn{6}{c}{\textbf{Prompt Alignment}}   
                                      \\ \cline{2-7} 

\textbf{DALL-E 3} & 1012 & 961 & \textbf{1055} & 1021 & 922 & \textbf{1113} \\
\textbf{Midjourney v6.1} & 971 & 1012 & 952 & 974 & 1033 & 899 \\
\textbf{Stable Diffusion 3 Medium} & \textbf{1021} & \textbf{1030} & 995 & \textbf{1028} & \textbf{1059} & 993 \\
\textbf{FLUX.1-dev} & 1003 & 1005 & 998 & 1004 & 1011 & 996 \\
\textbf{Playground v2.5} & 992 & 993 & - & 973 & 974 & - \\
\bottomrule
\end{tabular}
}
\caption{(MLE-)Elo ratings per model for the three separate criteria for all, English (en) and German (de) battles; The model Playground v2.5 was disabled for German language prompts.}
\label{tab:bafismleelobias}
\end{table}

\textbf{Image Quality:} Next, we focus on votes regarding image quality and investigate the (MLE-)Elo ratings of the models (see Tab. \ref{tab:bafismleelobias}). The evaluation indicates that the FLUX.1-dev model has the highest rating across all categories, while Midjourney v6.1 consistently holds 2nd place. The SD 3 Medium and Playground v2.5 models rank 3rd and 4th, respectively, with DALL-E 3 consistently last.

\textbf{Prompt Alignment:} Finally, we analyze the votes related to the prompt alignment criterion, as shown in Tab. \ref{tab:bafismleelobias}. The results indicate that DALL-E 3 in German and SD 3 Medium in English achieve the highest ratings, with SD 3 Medium consistently in first place ahead of DALL-E 3. Notably, DALL-E 3 drops to last place in English, while Midjourney v6.1 ranks second; however, it is consistently last in German battles. The FLUX.1-dev model holds third place, while Playground v2.5 fluctuates between 4th and 5th.

While Elo ratings align with the dataset analysis for bias and prompt alignment, DALL·E 3 is downvoted for perceived image quality compared to FLUX.1-dev. We assume that FLUX.1-dev exhibited greater variability in generating photorealistic images.
\section{Discussion}
\label{sec:discussion}

Here, we discuss the main findings from our analysis. 


\textbf{Bias:} With respect to MAD scores (see Fig. \ref{fig:bias_mad_gender_ethnicity} (a)) indicate a bias favoring males in the examined models, with only DALL-E 3 achieving nearly optimal results regarding gender bias, while FLUX.1-dev is the first model to show a preference for female faces. Additionally, the analysis of ethnicity (see Fig. \ref{fig:bias_mad_gender_ethnicity} (b)) reveals a bias toward faces of white ethnicity, which is more pronounced for German prompts compared to English ones. Notably, when focusing on English-language battles, Midjourney v6.1 has the highest score in terms of human preference (see Tab. \ref{tab:bafismleelobias}), contrasting with the statistical evaluation where DALL-E 3 consistently outperformed it in gender (see Fig. \ref{fig:bias_gender}) and ethnicity distributions \ref{tab:ethnicity_distribution}).

Furthermore, FLUX.1-dev and Playground v2.5 consistently rank among the lowest in the (MLE-)Elo ratings for bias in both languages. The statistical evaluation revealed that FLUX.1-dev shows a bias for females in English prompts, but this preference did not positively impact its (MLE-)Elo scores. This suggests that the direction of gender bias has no significant influence w.r.t. human preference. Lastly, DALL-E 3 demonstrated a clear correlation between its statistical evaluation results and (MLE-)Elo scores, consistently achieving above-average results in gender and ethnicity distributions, which aligns with our expectations.

\textbf{Image Quality:} The human preference data (see Tab. \ref{tab:bafismleelobias}) shows that FLUX.1-dev achieves the highest (MLE-)Elo scores, while DALL-E 3 ranks best in FID evaluations, and FLUX.1-dev coming in third (see Tab. \ref{tab:fidscores}). This discrepancy raises questions about the FID metric's reflection of human preferences, especially since DALL-E 3 consistently ranks last in human evaluations. The MagFace metric, which measures FIQ for recognition, indicates that SD 3 Medium scores highest, followed by Playground v2.5 and DALL-E 3, while FLUX.1-dev and Midjourney v6.1, which excel in (MLE-)Elo ratings, receive the lowest scores.

These contradictions suggest that FID and MagFace metrics may not capture important factors influencing human preferences. To explore this, we compared (MLE-)Elo rankings with those from two other platforms: \href{https://imgsys.org/}{IMGSYS by fal} and \href{https://artificialanalysis.ai/text-to-image/arena}{T2I Arena by Artificial Analysis} (see Tab. \ref{tab:bafiseloextern}). Notably, these platforms are not open-source and focus on overall image quality without specific evaluations for criteria like image quality. A correlation exists between BAFIS preference rankings and those from the other platforms, although DALL-E 3 and Midjourney v6.1 are missing from IMGSYS. The T2I Arena generally aligns with our rankings, except for SD 3 Medium, which performs better in BAFIS. Overall, these findings support the hypothesis that established metrics like FID and MagFace do not adequately represent human preferences regarding synthetic image quality.

\begin{table}[]
\centering
\resizebox{\columnwidth}{!}{%
\begin{tabular}{@{}lcccccc@{}}
\toprule
\textbf{Model} & Elo (fal) & Rang (fal) & Elo (arificalanalysis) & Rank (arificalanalysis)\\
\midrule
\textbf{FLUX.1-dev} & \textbf{1161} & \textbf{2} & \textbf{1108} & \textbf{3} \\
\textbf{Midjourney v6.1} & - & - & 1105 & 4 \\
\textbf{Playground v2.5} & 1079 & 6 & 1047 & 7 \\
\textbf{DALL-E 3} & - & - & 1026 & 9 \\
\textbf{Stable Diffusion 3 Medium} & 1006 & 19 & 992 & 12 \\
\bottomrule
\end{tabular}
}
\caption{Elo scores per model from the evaluation platforms: IMGSYS by fal and T2I Arena by Artificial Analysis \cite{falai_imgsysorg_nodate} and \cite{aa_arena_nodate}.}
\label{tab:bafiseloextern}
\end{table}

\textbf{Prompt Alignment:} Our preference data (see Tab. \ref{tab:bafismleelobias}) shows that DALL-E 3 in German and SD 3 Medium in English achieve the highest Elo ratings. While Playground v2.5 scores the highest in prompt alignment for English prompts using CLIP \cite{radford2021learningtransferablevisualmodels}, this is not reflected in the (MLE-)Elo ratings, where it consistently ranks 4th. This discrepancy raises questions about whether CLIP's similarity scores adequately capture prompt alignment in relation to human preferences.

DALL-E 3 emerges as an outlier, demonstrating high similarity between German "direct" images and English "direct" prompts, potentially due to its prompt revision feature. The "prompt revision" is a feature from DALL-E3, which automatically rewrites transferred prompts for security reasons. This resulted in more detailed prompts that can enhance image quality and also translates German prompts into English revisions to furhter improve textual alignment.

The (MLE-)Elo ratings for German battles also support this, indicating a significant positive influence of prompt revision on human ratings. Conversely, while SD 3 Medium ranks first for English-language battles and overall evaluations (see Tab. \ref{tab:bafiseloextern}), its results do not correlate with the prompt alignment evaluations from BAFIS or the Elo ratings from the \href{https://imgsys.org/}{IMGSYS by fal} and \href{https://artificialanalysis.ai/text-to-image/arena}{T2I Arena by Artificial Analysis} platforms. This suggests that BAFIS's focus on occupation-related portraits may be a key factor influencing human preferences regarding prompt alignment. Finally, we would like to emphasize that, although human feedback is valuable, it can also introduce biases.
\section{Conclusion}
\label{sec:conclusion}

We presented BAFIS, a multilingual benchmark for examining socio-economic bias and evaluating prompt alignment and image quality in T2I models. Building on the MAGBIG dataset, BAFIS adds 151 German and 151 English group prompts, with potential for future language expansions. We provided a dataset of 21,140 synthetic images generated from 1,057 prompts using five advanced T2I models, serving as a valuable tool for bias investigation and T2I model performance assessment based on human preferences.

Our evaluation found that models exhibit gender-specific and ethnic biases, with stronger biases in German prompts, favoring faces of white ethnicity. DALL-E 3 received the highest prompt alignment ratings in German battles, significantly correlated with CLIP values, indicating that prompt revisions enhance alignment with human preferences. However, it is rated lower in image quality and alignment for English battles. These findings highlight the importance of understanding human preferences in image generation and encourage further research on benchmarks assessing bias in T2I models. 

As actionable recommendations, we urge researchers and developers to establish guidelines addressing bias in T2I models, to ensure datasets are diverse and representative, and to conduct regular assessments of model bias. Engaging with policymakers is crucial for shaping regulations that promote ethical AI practices, and establishing feedback channels for users can help identify bias-related concerns.


{
    \small
    \bibliographystyle{ieeenat_fullname}
    \bibliography{main}
}

\end{document}